\documentclass[12pt,a4]{amsart}
\usepackage{latexsym}
\usepackage{amsmath}
\usepackage{amssymb}
\usepackage{amsthm}
\usepackage{amscd}
\usepackage{color}
\definecolor{dblue}{rgb}{0,0,0.45}
\definecolor{red}{rgb}{0.7,0,0}
\usepackage[notref,notcite]{showkeys} 

\usepackage{multicol}
\usepackage{color}

\numberwithin{equation}{section}

\newtheorem{theorem}{Theorem}[section]
\newtheorem{lemma}[theorem]{Lemma}

\theoremstyle{definition}

\newtheorem{remark}[theorem]{Remark}
\newtheorem{definition}[theorem]{Definition}

\theoremstyle{remark}

\title{A global universality of two-layer neural networks with ReLU activations}

\author{Naoya Hatano, Masahiro Ikeda, Isao Ishikawa, and Yoshihiro Sawano}
\address[Naoya Hatano]{Department of Mathematics, Faculty of Science and Technology, Keio University, 3-14-1 Hiyoshi, Kohoku-ku, Yokohama 223-8522, Japan/Center for Advanced Intelligence Project, RIKEN, Japan, and Department of Mathematics, Chuo University, 1-13-27, Kasuga, Bunkyo-ku, Tokyo 112-8551, Japan,}
\address[Masahiro Ikeda]{Center for Advanced Intelligence Project, RIKEN, Japan/Department of Mathematics, Faculty of Science and Technology, Keio University, 3-14-1 Hiyoshi, Kohoku-ku, Yokohama 223-8522, Japan}
\address[Isao Ishikawa]{Center for Advanced Intelligence Project, RIKEN, Japan, and Department of Engineering for Production and Environment, Graduate School of Science and Engineering, Ehime University, 3 Bunkyo-cho, Matsuyama, Ehime 790-8577, Japan,}
\address[Yoshihiro Sawano]{Department of Mathematics, Faculty of Science and Technology, Keio University, 3-14-1 Hiyoshi, Kohoku-ku, Yokohama 223-8522, Japan/Center for Advanced Intelligence Project, RIKEN, Japan, and Department of Mathematics, Chuo University, 1-13-27, Kasuga, Bunkyo-ku, Tokyo 112-8551, Japan}

\email[Naoya Hatano]{n.hatano.chuo@gmail.com}
\email[Masahiro Ikeda]{masahiro.ikeda@riken.jp}
\email[Isao Ishikawa]{ishikawa.isao.zx@ehime-u.ac.jp}
\email[Yoshihiro Sawano]{yoshihiro-sawano@celery.ocn.ne.jp}

\begin{document}
\maketitle
\begin{abstract}
In the present study, we investigate a universality of neural networks, which concerns a density of the set of  two-layer neural networks in a function spaces. 
There are many works
that handle the convergence
over compact sets.
In the present paper,
we consider a global convergence
by introducing a norm suitably,
so that our results will be uniform
over any compact set.
\end{abstract}

\section{Introduction}
%
Neural network is a function that models a neuron system of a biological brain, and defined as alternate compositions of affine map and a nonlinear map.
The nonlinear map in a neural network is called activation function.  
The neural networks have been
playing
a central role in the field of machine learning with a vast number of applications in the real world
in the last decade.

We focus on a two-layer feed-forward neural network with ReLU
 (Rectified Linear Unit)
 activation, that is a function $f:\mathbb{R}\rightarrow\mathbb{R}$ in the form of $f(x) = \sum\limits_{i=1}^r c_i {\rm ReLU}(a_ix+b_i)$ for some $a_1,b_1,c_1,\dots,a_r,b_r,c_r\in \mathbb{R}$.
Here, the function ${\rm ReLU}$ is called the rectified linear unit defined by 
\[{\rm ReLU}(x):=\max(x,0).\]
The ReLU is one of the most popular activation functions for feed-forward neural networks in practical machine learning tasks for real world problems.

We consider the space of two-layer feedforward neural networks defined by the following linear space
\[
{\mathcal X}
:=
{\rm Span}\left(
\left\{
{\rm ReLU}(a \cdot+b)\,:\,a \ne 0, b \in {\mathbb R}
\right\}\right).
\]
Then, it is natural to ask ourselves whether
${\mathcal X}$ spans a dense subspace
of a function space (topological linear space).
Historically, the density property of $\mathcal{X}$ in the space ${\rm C}(\mathbb{R})$ of continuous functions on $\mathbb{R}$ is investigated by several authors (\cite{Cybenko89, Funahashi89, HSW89})
as it is important to finding a feed-forward neural network $f\in \mathcal{X}$ that approximates an unknown continuous function.
Here, the topology of ${\rm C}(\mathbb{R})$ is 
generated
by the seminorms $h\mapsto \sup_{x\in K}|h(x)|$,
where $K$ ranges over all compact sets in $\mathbb{R}$.
Thus, the approximation property of two-layer feed-forward neural networks makes sense only on a local domain.

In this study, we prove a approximation property of $\mathcal{X}$ in a global sense.
More precisely, we prove the space $\mathcal{X}$ is dense in the Banach subspace of ${\rm C}(\mathbb{R})$ defined as
\[
{\mathcal Y}
:=
\left\{f\in{\rm C}({\mathbb R})\,:\,
\lim_{x \to \pm\infty}\frac{f(x)}{1+|x|}
\mbox{ exists }\right\}
\]
equipped with the norm
\[
\|f\|_{{\mathcal Y}}
:=
\sup_{x \in {\mathbb R}}\frac{|f(x)|}{1+|x|}.
\]
Note that any element in ${\mathcal Y}$, divided by $1+|\cdot|$,
is a continous funciton over 
$\overline{\mathbb R}:={\mathbb R} \cup \{\pm \infty\}$.
Our main result in this paper is as follows:
\begin{theorem}\label{thm:201113-1}
The linear subspace
${\mathcal X}$ is dense
in
${\mathcal Y}$.
\end{theorem}
Before we conclude this section,
we will offer some words on some existing results.
See
\cite{SoMu17}
for the $L^2$-approximation
over the real line.
Other attempts has been made
to grasp the neural network
by the use of the Radon transform
\cite{CaDi89}
or 
by considering some other topologies
\cite{HSW89,SGT20}.

\section{Proof of the main theorem}
\begin{definition}
We define a linear operator $A:f \in {\mathcal Y} \mapsto \frac{f}{1+|\cdot|} \in {\rm BC}(\overline{\mathbb R})$.
\end{definition}

\begin{lemma}
The operator $A:\mathcal{Y}\rightarrow {\rm BC}(\overline{\mathbb{R}})$ is an isomorphism
from ${\mathcal Y}$ to ${\rm BC}(\overline{\mathbb R})$.
\end{lemma}

A tacit understanding here is that
we extend $\frac{f}{1+|\cdot|}$,
which is initially defined over ${\mathbb R}$,
continuously to $\overline{\mathbb R}$.

Thus, 
any continuous functional
on ${\mathcal Y}$ is realized by a Borel measure
over $\overline{\mathbb R}$.

Our theorem can recapture the case
where the underlying domain is bounded.
Indeed, if the domain $\Omega$ is contained
in $[-R,R]$ for some $R>0$,
then we have
\[
\|f\|_{L^\infty(\Omega)}
\le (1+R)
\|f\|_{{\mathcal Y}}
\quad (f \in {\mathcal Y}),
\]
which will give results
by Cybenko
\cite{Cybenko89}
and Funahashi
\cite{Funahashi89}.

Now we start the proof of Theorem \ref{thm:201113-1}.
As Cybenko did in \cite{Cybenko89},
take any measure
$\mu$
over $\overline{\mathbb R}$
such that $\mu$ annihilates
${\mathcal X}$.
We will show that $\mu=0$.
Once this is proved,
from 
the Riesz representation theorem
we conclude that the only linear functional
that vanishes on ${\mathcal X}$
is zero.
Using
the Hahn-Banach theorem,
we see that
${\mathcal X}$ is dense in ${\mathcal Y}$.

Remark that
\[
\max(1-|x-1|,0)
=
{\rm ReLU}(x)+{\rm ReLU}(x-2)
-2
{\rm ReLU}(x-1)
\quad (x \in {\mathbb R}).
\]
Thus,
any element in 
$C_{\rm c}({\mathbb R})$
can be approximated by a function
${\mathcal X}$
in the $L^\infty$-norm.
Since $\mu$ annihilates
$C_{\rm c}({\mathbb R})$,
it follows that
$\mu$ is not supported on ${\mathbb R}$.
Recall that 
$\mu|_{\mathbb R}=0$,
so that
$\mu$ is supported on $\pm\infty$.
It remains to show that
$\mu(\{\pm\infty\})=0$.
Consider
\[
f(x)={\rm ReLU}(x)-{\rm ReLU}(x-1) \quad (x \in {\mathbb R}).
\]
Remark that
\[
0=\int_{\overline{\mathbb R}}f(x)d\mu(x)=\mu(\{\infty\}).
\]
Likewise if we test the condition on
$
g=f(-\cdot),
$
we obtain $\mu(\{-\infty\})=0$.

Thus,
we conclude
that
${\mathcal X}$ is dense
in
${\mathcal Y}$.

\begin{remark}
The set
$\{f,g\} \cup C_{\rm c}({\mathbb R})$ spans a dense
subspace in
${\mathcal Y}$, where $f$ and $g$ are functions given in the above proof.
\end{remark}

\vspace{10pt}
{\em Availability of data and material}.
No data and material were used to support this study.

\vspace{10pt}
{\em Competing interests}.
The authors declare that there are no conflicts of interest regarding the publication of this paper.

\vspace{10pt}
{\em Funding}.
This work was supported by a JST CREST Grant (Number JPMJCR1913, Japan).
This work was also supported by the RIKEN Junior Research Associate Program.
The second author 
was
supported by a Grant-in-Aid for Young Scientists Research 
(No.19K14581), Japan Society for the Promotion of Science.
The fourth author 
was
supported by a 
Grant-in-Aid for Scientific Research (C) (19K03546), 
Japan Society for the Promotion of Science.

\vspace{10pt}
{\em Authors' contributions}.
The four authors contributed equally to this paper.
All of them read the whole manuscript and approved the content of the paper.

\vspace{10pt}
{\em Acknowledgements}.
The authors are thankful to Professor Ken-ichi Bannai 
at Keio University for giving us a chance 
to consider the problem.

\end{document}